%% file: main.tex
\documentclass{vgtc}                          

\ifpdf
  \pdfoutput=1\relax                   
  \usepackage{graphicx}                
  \DeclareGraphicsExtensions{.pdf,.png,.jpg,.jpeg} 
\else
  \usepackage{graphicx}                
  \DeclareGraphicsExtensions{.eps}     
\fi%

\graphicspath{{figures/}{pictures/}{images/}{./}} 

\usepackage{microtype}                 
\PassOptionsToPackage{warn}{textcomp}  
\usepackage{textcomp}                  
\usepackage{mathptmx}                  
\usepackage{times}                     
\usepackage{cite}                      
\usepackage{tabu}                      
\usepackage{booktabs}                  
\usepackage{xspace}

\usepackage[utf8]{inputenc}
\usepackage{epsfig}
\usepackage{graphicx}
\usepackage{amsmath}
\usepackage{amssymb}
\usepackage{enumitem}
\usepackage{makecell}
\usepackage{color}
\usepackage{array,multirow}

\onlineid{1434}

\vgtccategory{Research}

\vgtcinsertpkg

\usepackage{caption}
\usepackage{subcaption}

\title{RGB-D-E: Event Camera Calibration for Fast 6-DOF Object Tracking}

\author{Etienne Dubeau\textsuperscript{1}\thanks{e-mail: etienne.dubeau.1@ulaval.ca} %
\and Mathieu Garon\textsuperscript{1}\thanks{e-mail: mathieu.garon.2@ulaval.ca} %
\and Benoit Debaque\textsuperscript{2} \thanks{e-mail: benoit.debaque@ca.thalesgroup.com} %
\and Raoul de Charette\textsuperscript{3}\thanks{e-mail: raoul.de-charette@inria.fr} %
\and Jean-François Lalonde\textsuperscript{1}\thanks{e-mail: jean-francois.lalonde@gel.ulaval.ca} %
}

\affiliation{\scriptsize \textsuperscript{1}Université Laval  \quad\quad  \textsuperscript{2}Thales Digital Solutions \quad\quad  \textsuperscript{3}Inria}

\abstract{
Augmented reality devices require multiple sensors to perform various tasks such as localization and tracking. Currently, popular cameras are mostly frame-based (e.g. RGB and Depth) which impose a high data bandwidth and power usage. With the necessity for low power and more responsive augmented reality systems, using solely frame-based sensors imposes limits to the various algorithms that needs high frequency data from the environement. As such, event-based sensors have become increasingly popular due to their low power, bandwidth and latency, as well as their very high frequency data acquisition capabilities. In this paper, we propose, for the first time, to use an event-based camera to increase the speed of 3D object tracking in 6 degrees of freedom. This application requires handling very high object speed to convey compelling AR experiences. To this end, we propose a new system which combines a recent RGB-D sensor (Kinect Azure) with an event camera (DAVIS346). We develop a deep learning approach, which combines an existing RGB-D network along with a novel event-based network in a cascade fashion, and demonstrate that our approach significantly improves the robustness of a state-of-the-art frame-based 6-DOF object tracker using our RGB-D-E pipeline. Our code and our RGB-D-E evaluation dataset are available at \url{https://github.com/lvsn/rgbde_tracking}.
} 

\CCScatlist{
  \CCScatTwelve{Event camera}{Calibration}{6-DOF Object tracking}{Augmented reality};
}

\teaser{
\centering
  \setlength{\tabcolsep}{1pt}
  \begin{tabular}{cccc}
        \includegraphics[height=3.35cm]{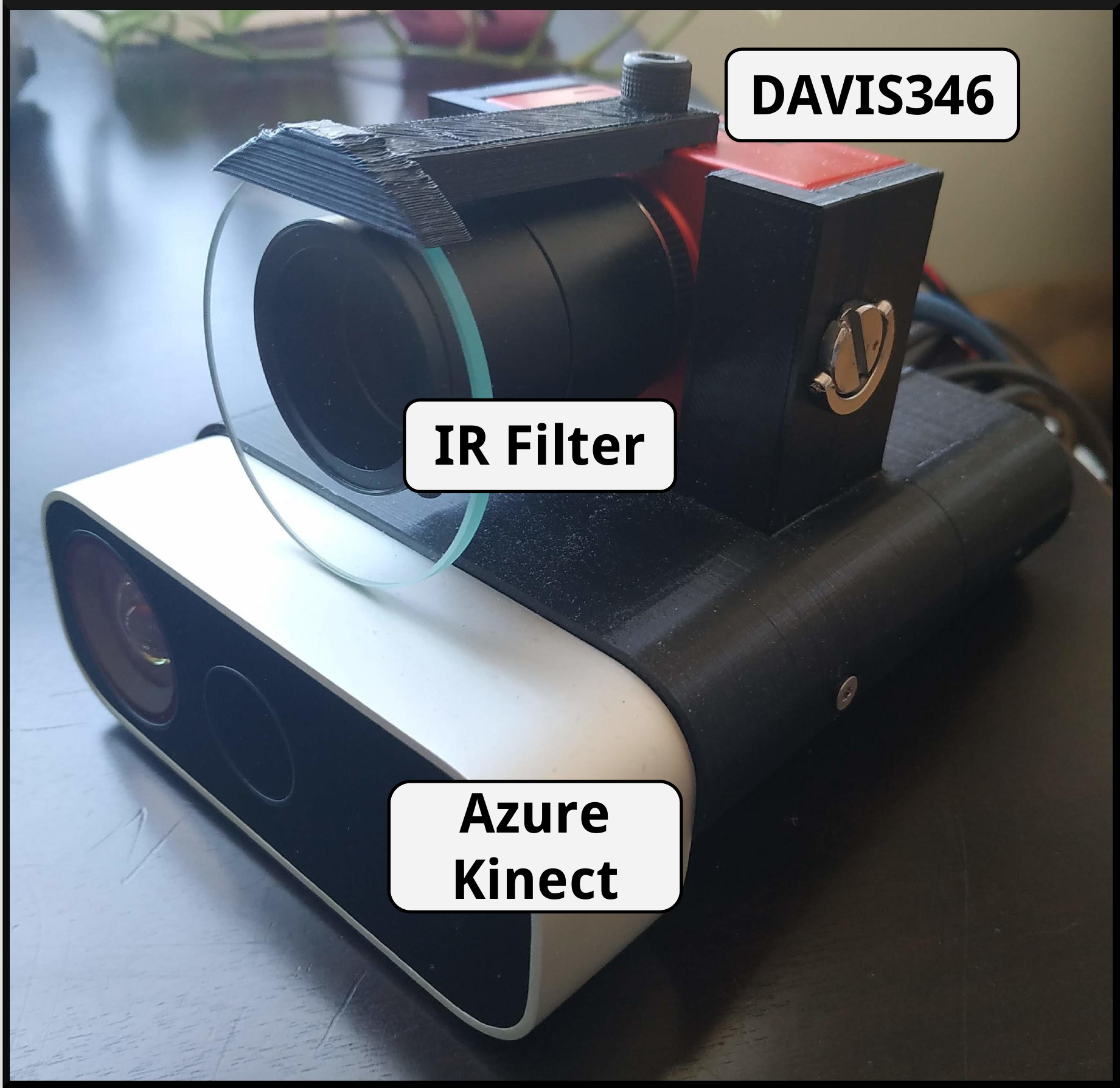} &
        \includegraphics[height=3.35cm]{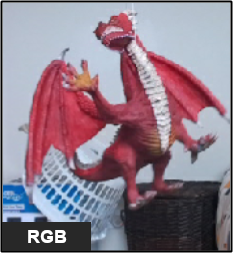} &
        \includegraphics[height=3.35cm]{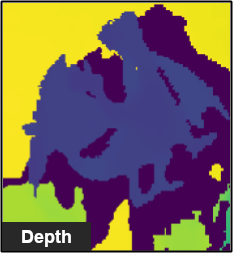} &
        \includegraphics[height=3.35cm]{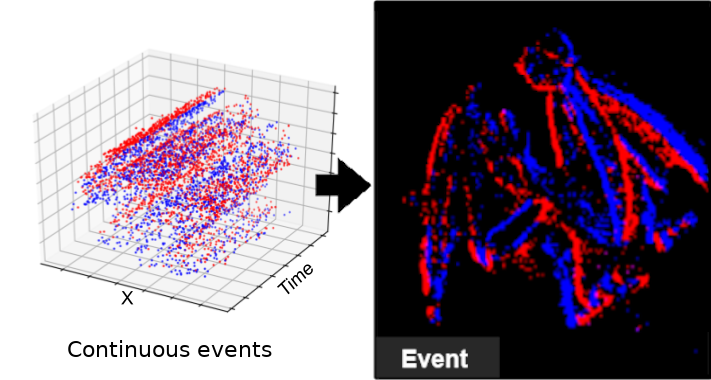} \\
  \end{tabular}
  \caption{Our RGB-D-E hardware setup uses a Kinect Azure (RGB-D) and a DAVIS346 event-based camera (E) for continuous events which are temporally binned. RGB-D-E data streams are spatially and temporally calibrated and used for 6 degree of freedom object tracking.}
  \label{fig:teaser}
 }

\begin{document}


\maketitle

\input{section/01_Introduction}

\input{section/02_RelatedWork}
\input{section/03_HardwareSetupandCalibration}

\input{section/04_Methodology}
\input{section/05_Evaluation}
\input{section/06_Discussion}

\acknowledgments{The authors wish to thank Jérémie Roy for his help with data acquision. This work was partially supported by a FRQ-NT Samuel de Champlain grant, the NSERC CRDPJ 524235 - 18 grant, and Thales. We thank Nvidia for the donation of the GPUs used in this research.}

\bibliographystyle{abbrv-doi}

\bibliography{biblio}
\end{document}

%% file: section/01_Introduction.tex
\section{Introduction}

Compelling augmented reality (AR) experiences are achieved through the successful execution of several tasks in parallel. Notably, simultaneous localization and mapping (SLAM)~\cite{newcombe2011kinectfusion}, hand tracking~\cite{mueller2017real}, and object tracking in 6 degrees of freedom (6-DOF)~\cite{garon2018framework} must all be executed efficiently and concurrently with minimal latency on portable, energy-efficient devices. 

This paper focuses on the task of 6-DOF rigid object tracking. In this scenario, successfully tracking the object at high speed is particularly important, since freely manipulating an object can easily result in translational and angular speeds of up to 1~m/s and $360^\circ/\mathrm{s}$ respectively. Despite recent progress on real time 6-DOF object tracking at 30~fps~\cite{garon2018framework,manhardt2018deep,li2018deepim}, these methods still have trouble with very high object motion and tracking failures are still common. Increasing the speed of 6-DOF object trackers is of paramount importance to bring this problem closer to real-world applications. 

To increase the speed of object tracking, one can trivially employ cameras with frame rates higher than 30~fps. Indeed, 90 and even 120~fps off-the-shelf cameras are available and could be used as a drop-in replacement. However, this comes at significant practical disadvantages: higher data bandwidth, increased power consumption (since the algorithms must be executed more often), and the necessity to have sufficient light in the scene since exposure times for each frame is necessarily decreased. 

In this work, we propose a system to increase the speed of 6-DOF object tracking applications with a minimal increase in bandwidth and power consumption. Specifically, we propose to \emph{combine} an event camera (specifically, the DAVIS346 camera) with an RGB-D camera (the Kinect Azure) into a single ``RGB-D-E'' capture system. The event camera offers several key advantages: very low latency ($20~\mu\mathrm{s}$), bandwidth, and power consumption (10--30~mW), all while having much greater dynamic range (120~dB vs 60~dB) than frame-based cameras. 

This paper makes the following contributions. First, we show how to calibrate the setup both spatially and temporally. Second, we provide a new challenging publicly available 6-DOF evaluation dataset that contains approximately 2,500 RGB-D-E frames of a real-world object with high-speed motion with the corresponding ground truth pose at each frame. Third, we propose what we believe to be the first 6-DOF object tracker that uses event-based data. Similar to previous work~\cite{garon2018framework,manhardt2018deep,li2018deepim}, our approach assumes that the object to track must be rigid (non-deforming) and its textured 3D model must be known a priori. Finally, we demonstrate through a quantitative analysis on our real evaluation dataset that, using an extension of an existing deep learning approach for 6-DOF object tracking results in a threefold decrease in the number of tracking failures and achieves robust tracking results on fast free interaction motions. We believe this paper brings 6-DOF object tracking one step closer to real-world augmented reality consumer applications.



%% file: section/02_RelatedWork.tex
\section{Related work}

The majority of computer vision systems rely on established frame-based camera architectures, where the scene irradiance is captured synchronously at each pixel or in a rapid, rolling shutter sequence~\cite{liang2008analysis}. However, such cameras need to stream large amount of data (most of which redundant), making them power- and bandwidth-hungry. Recently, a newer camera architecture with a event-based paradigm~\cite{lichtsteiner2008128} is gaining popularity. By triggering events on each pixel asynchronously when the brightness at that pixel changes by certain threshold, event-based camera can stream at a much higher frequency while consuming less power. A branch of computer vision research now focuses on developing algorithms to take advantage of this new type of data.

\textbf{Event-based applications.} Event-based sensors bring great promises in the field as their low power consumption makes them ideal for embedded systems such as virtual reality headset~\cite{gallego2017event}, drones~\cite{vidal2017hybrid,delmerico2019we} or autonomous driving~\cite{maqueda2018event}. 
Their high-speed resolution also enables the design of robust high-frequency algorithms like SLAM~\cite{nguyen2019real,vidal2018ultimate,weikersdorfer2014event,rebecq2016evo,bryner2019event,gallego2017event,kim2016real} or fast 2D object tracking~\cite{glover2017robust,mitrokhin2018event}. While related to our work since we also focus on tracking, all related works are still restricted to tracking objects in the 2D image plane. 
In this paper, we extend the use of event cameras to the challenging task of fast 6-DOF object tracking by building over a state-of-the-art frame-based 6-DOF object tracker~\cite{garon-tvcg-17}. Different from other works, we benefit from RGB, Depth and Event data to propose the first RGB-D-E 6-DOF object tracker. 

\textbf{Deep learning with events.} Using event-based data is not straightforward since the most efficient deep architectures for vision are designed for processing conventional image data (e.g. CNNs). In fact, it is still unclear how event-based data should be provided to networks since each event is a 4-dimensional vector storing time, 2D position, and event polarity. Experimental architectures such as spiking neural networks~\cite{maass2004computational} holds great promises but are currently unstable or difficult to train~\cite{lee2016training}. With conventional deep frameworks, events can be converted to 2D tensors only by discarding both time and polarity dimensions \cite{rebecq2017real} or to 3D tensors by discarding either of the two dimensions \cite{maqueda2018event,zhu2019unsupervised}. Recent work~\cite{rebecq2019events,scheerlinck2020fast} has demonstrated that conventional grayscale frames can be reconstructed from event data, opening the way to the use of existing algorithms on these ``generated'' images. In this paper, we favor the Event Spike Tensor formulation from Gehrig et al.~\cite{Gehrig_2019}, where time dimension is binned. This allows us to exploit event data directly without requiring the synthesis of intermediate images, while maintaining a fast convolutional network architecture. 

\textbf{Event-based datasets.} Finally, large amount of training data is required. While a few events datasets exist mostly for localization/odometry~\cite{delmerico2019we,li2018interiornet,mueggler2017event} or 2D object tracking~\cite{hu2016dvs}, there are, as of yet, no 6-DOF object tracking dataset which contains event data. Instead, event data can be synthesized with a simulator such as~\cite{Rebecq18corl} which allows various types of data augmentation~\cite{rebecq2019high}. Our experiments show that a network can be trained without using real data and is not critically affected by the real-synthetic domain gap.








%% file: section/03_HardwareSetupandCalibration.tex
\section{System overview and calibration}

\begin{figure}
     \centering
     \setlength{\tabcolsep}{1pt}
     \begin{tabular}{cc}
        \includegraphics[width=0.5\linewidth]{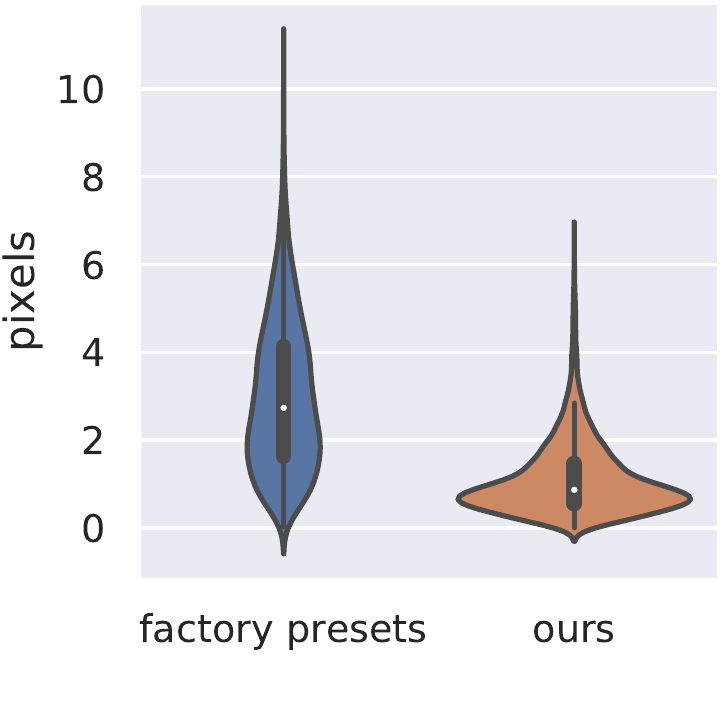} &
        \includegraphics[width=0.5\linewidth]{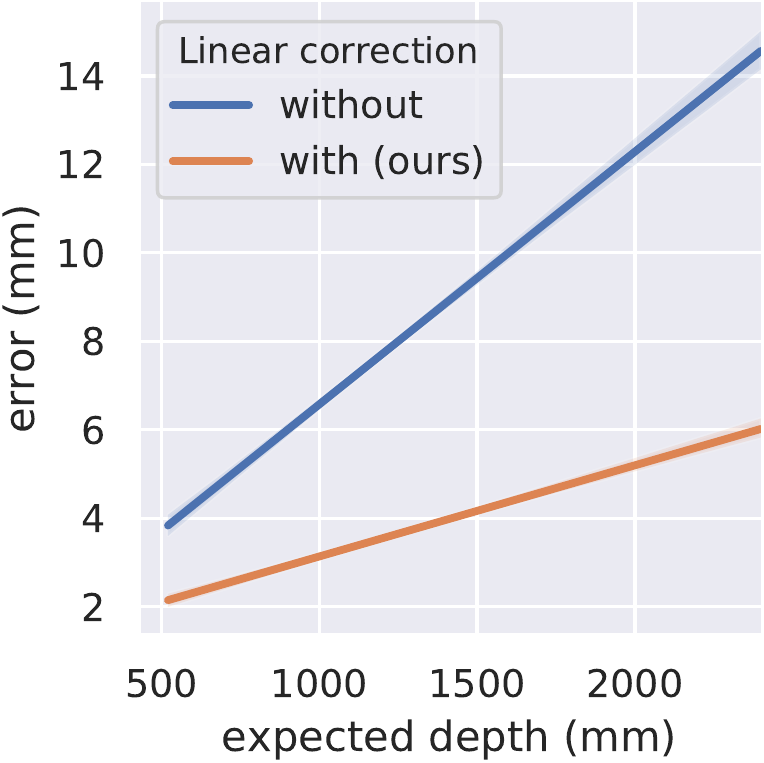} \\
        (a) & (b)
     \end{tabular}
    
    \caption{Comparison between the factory presets and our calibration. (a)~The reprojection error from the Depth to RGB image ($\mathbf{T}_\mathrm{Depth}^\mathrm{RGB}$) computed on 51 matching planar checkerboard images. (b)~Linear regression of the Kinect depth map error compared to the expected depth, computed on calibration target corners.}
    \label{fig:calibration_k4a}
\end{figure}

In this section, we describe our novel RGB-D-E hardware setup, which combines a Microsoft Kinect Azure (RGB-D) with a DAVIS346 event camera (E).

\subsection{System overview}

As illustrated in Fig.~\ref{fig:teaser}, the DAVIS346 event camera is rigidly mounted over the Kinect Azure using a custom-designed, 3D-printed mount. We observed that the modulated IR signal projected on the scene by the Time-of-Flight (ToF) sensor in the Kinect triggered multiple events in the DAVIS346 camera. To remedy this limitation, an infrared filter is placed in front of the event camera lens. 

\begin{figure}[!t]
	\centering
	\begin{tabular}{cc}
		\includegraphics[width=0.45\linewidth]{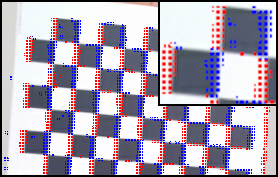} &
		\includegraphics[width=0.45\linewidth]{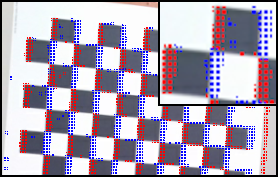} \\
		Frame $t_0$ & Frame $t_0 + \Delta t$ \\
		Event $[t_0, t_0 + \Delta t]$ & Event $[t_0, t_0 + \Delta t]$
	\end{tabular}
	\caption{Events from the DAVIS346 camera projected on Kinect RGB frames on a moving calibration target. The projection is obtained using the calibrated transformation $\mathbf{T}_\mathrm{Event}^\mathrm{RGB}$. 
		Events of a moving checkerboard are accumulated and represented as red (negative) and blue (positive) on both frames. Pixels with more than one event are presented to reduce distraction by noise. Frames are captured at a $\Delta t = 1/15~\mathrm{s}$ interval and events between $t_0$ and $t_0 + \Delta t$. A proper alignment of events with the checkerboard demonstrates that the system is calibrated both spatially and temporally. }
	\label{fig:transfo}
\end{figure}

\subsection{Spatial calibration}
\label{sec:spatial_calibration}

Our system contains 3 cameras that must be calibrated: the Kinect RGB, the Kinect Depth and the DAVIS346 sensor. In this paper, we describe a coordinate system transformation with the notation $\mathbf{T}_a^b$, denoting a transformation matrix from coordinate frame $a$ to $b$. 

The intrinsic parameters of each camera can be computed with a standard method \cite{zhang2000flexible}. The checkerboard corners can easily be found using the color frame and the IR image from the Kinect Azure. Calibrating an event-based sensor is usually more difficult, however the DAVIS346 possesses an APS sensor (gray scale frame-based capture) that is spatially aligned with the event-based capture sensor. We thus use the APS sensor to detect the target corners that will be used for the intrinsic and extrinsic calibration.

\textbf{Intrinsics.} We capture images where a checkerboard target ($9 \times 14$ with 54~mm squares) is positioned in a spatial uniform distribution in the frustum of each camera. To account for varying fields of view, 199 images were captured for the Kinect RGB, 112 for the Kinect Depth, and 50 for the DAVID346. 
For each sensor, we retrieve the intrinsic parameters (focal and image center) with a lens distortion model including 6 radial and 2 tangential parameters. 

\textbf{Extrinsics.} We retrieve the rigid transformations $\mathbf{T}_\mathrm{RGB}^\mathrm{Depth}$ and $\mathbf{T}_\mathrm{Event}^\mathrm{Depth}$ by capturing images of the target in overlapping frustums. Once the 3D points are retrieved from the previously computed camera intrinsic and the known checkerboard geometry, PnP~\cite{fischler1981random} is used to retrieve the 6-DOF transformation between each camera.

Finally, we compare our calibration procedure with the factory presets of the Kinect Azure. Motivated by previous work~\cite{chen2018calibrate} that demonstrate lower accuracy errors with factories presets calibration we capture a \emph{test} dataset of 45 target images and show that we obtain a lower reprojection error in Fig.~\ref{fig:calibration_k4a}-(a).

\begin{figure*}[!t]
\centering
\footnotesize
\setlength{\tabcolsep}{1pt}
\begin{tabular}{ccccccc}

\rotatebox{90}{\hspace{2.5em}Real data} & 
\includegraphics[width=.15\linewidth]{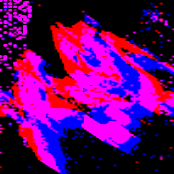} &
\includegraphics[width=.15\linewidth]{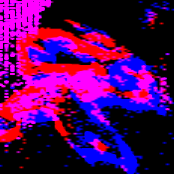} &
\includegraphics[width=.15\linewidth]{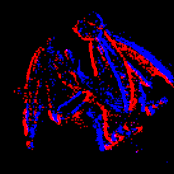} &
\includegraphics[width=.15\linewidth]{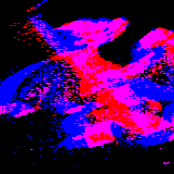} &
\includegraphics[width=.15\linewidth]{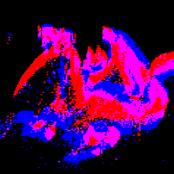} &
\includegraphics[width=.15\linewidth]{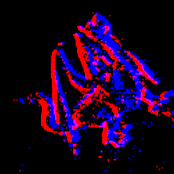} \\
\rotatebox{90}{\hspace{2.2em}Synthetic data} & 
\includegraphics[width=.15\linewidth]{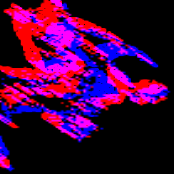} &
\includegraphics[width=.15\linewidth]{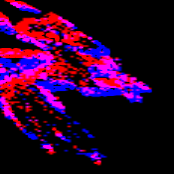} &
\includegraphics[width=.15\linewidth]{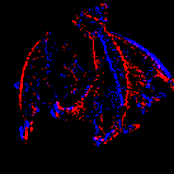} &
\includegraphics[width=.15\linewidth]{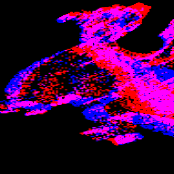} &
\includegraphics[width=.15\linewidth]{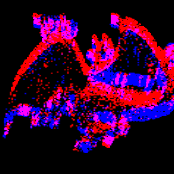} &
\includegraphics[width=.15\linewidth]{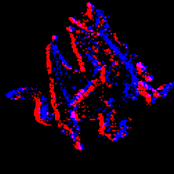} \\
\end{tabular}
\caption{Qualitative comparison between real (top) and synthetic events (bottom). Synthetic frames are generated with the event simulator of Rebecq et al.~\cite{Rebecq18corl}, where the pose of the synthetic object is adjusted to match the real. Event polarities are displayed as blue (positive) and red (negative).}
\label{fig:simul_compare}
\end{figure*}

\subsection{Depth correction}

As \cite{hodan2017t,garon2018framework} reported for the Kinect 2, we also found that the depth from the Kinect Azure has an offset that changes linearly w.r.t the depth distance and average in an error in the range of 8.5~mm. We compare the target points from the calibration dataset with the depth pixels in each frame and fit a 2nd-degree polynomial to the errors w.r.t to their distance to the camera. In Fig.~\ref{fig:calibration_k4a}b, we show the error with and without the polynomial correction on the test calibration set. Using the correction, the mean error on the test calibration set is less than 4~mm.

\subsection{Temporal synchronization}


In a multi-sensors setup, each sensor acquires data at its own frequency aligned with its inner clock. For time-critical applications, such as fast object tracking, it is required to synchronize the sensors clocks to ensure temporal alignment of the data. 
Technically, this is commonly addressed with synchronization pulses emitted by a \textit{master} sensor at the beginning of each data frame acquisition, subsequently triggering the acquisition of other \textit{slave} sensors.

In our setup, both Kinect and DAVIS346 support hardware synchronization but we found that the Kinect (master) emits a variable number of pulses before the first RGB-D frame. This led to incorrect triggering of DAVIS346 (slave) and thus temporal misalignment of RGB-D and Event data.
Because pulses are always emitted at the same frequency, we fix this by computing the pulses offset $\delta$ as
\begin{equation}
\delta = \lfloor \mathrm{RGBD}^t_0 \times \mathrm{RGBD}^{\mathrm{fps}} \rfloor \,,
\end{equation}
where $\mathrm{RGBD}^t_0$ is the timestamp of the first RGB-D frame and $\mathrm{RGBD}^{\mathrm{fps}}$ is the Kinect framerate (here, 30). Following this, we can pair RGBD and Event frames as $\big(\mathrm{RGBD}_i, \mathrm{E}_{i+\delta}\big)$.
Fig.~\ref{fig:transfo} illustrates the projection of events captured on a moving checkerboard. The events are captured between the two RGB frames. Alignment with the borders of the pattern shows the temporal and spatial calibration.

%% file: section/04_Methodology.tex

\section{Fast 6-DOF object tracking}

With the sensors spatio-temporally calibrated, we enhance an existing tracking framework by the addition of the new event modality (E). We build on the work of Garon et al.~\cite{garon-tvcg-17,garon2018framework} who propose a deep learning approach of robust 6-DOF object tracking, which relies on the refinement between a render of the object at the current pose estimate and the current Kinect RGB-D frame. While this method is robust to occlusion and small displacements, we notice that it is significantly impacted by larger motions (over 0.5 m/s), possibly because of the motion blur induced. Additionally, the network in~\cite{garon2018framework} is fundamentally limited by a maximum pose translation of 2 cm between two frames. We note that increasing the sensor frame rate is also not a practical solution as the network computation time is the main bottleneck. In this section, we improve the tracker reactivity and robustness with the addition of an event-specific network. In the following, we first describe the generation of synthetic data for training and proceed to explain how frame-based and event-based trackers are jointly used.

\subsection{Training data generation}

Despite the existence of event datasets \cite{hu2016dvs,li2018interiornet,zhu2019eventgan}, none of them provide event data with 6-DOF object pose. Since capturing a dataset of sufficient magnitude and variety for training a deep network is prohibitive, we rely on synthetic data generated from an event camera simulator~\cite{Rebecq18corl}. The engine renders a stream of events that represent changes in pixel brightness, thus mimicking event-based sensors.
We build a training dataset by generating sequences of events where our target object (here, a toy dragon) is moved in front of a static camera. 
We acquire a textured 3D model of the dragon with a Creaform GoScan\texttrademark\xspace handheld 3D scanner at 1~mm voxel resolution, subsequently cleaned manually using Creaform VxElements\texttrademark\xspace to remove background and spurious vertices. 
As the camera remains stationary, we simulate the scene background with a random RGB texture from the SUN3D dataset~\cite{xiao2013sun3d} applied on a plane orthogonal to the virtual camera optical axis. We next describe the simulation setup followed by various data augmentation strategies applied to the data sample. 
\\

\textbf{Simulation details.} Event sequences are generated by first positioning the object in front of the camera at a random distance $d \sim \text{U}(0.45~\mathrm{m}, 0.8~\mathrm{m})$ (where $\text{U}(a, b)$ denotes a uniform distribution in the $[a, b]$ interval) and a random orientation. The center of mass of the object is aligned with the optical axis of the camera, so the object appears in the center of the frame. 
The object is then displaced by a random pose transformation over 33~ms and the generated events are recorded. The transformation is generated by first sampling two directions on the sphere using spherical coordinate $(\theta,\phi)$ with $\theta \sim \text{U}(-180^\circ, 180^\circ)$ and $\phi = \cos^{-1}(2x - 1)$, where $x \sim \text{U}(0, 1)$ as in~\cite{garon2018framework} and then sample the magnitude of the translation and rotation with $\text{U}(0~\text{m}, 0.04~\text{m})$ and $\text{U}(0^\circ, 35^\circ)$ respectively. A 3D bounding box of size 0.207~m around the object is projected on the image plane. The event spatial axes are then cropped according to the projected bounding box and resized with bilinear interpolation to a spatial resolution of $150\times150$. Each 33~ms pose transformation generates a set of $N$ events storing $\{t, x, y, p\}_{i=1..N}$ where $t$ is time, $x$ and $y$ are pixel coordinates and $p$ the polarity of the event (positive or negative, indicating a brighter or darker transition respectively). A total of 10 such event sets are simulated for each background image, leading to 180,000 training and 18,000 validation sets.
\\

\textbf{Data augmentation.} To maximize the reliability of our simulations, we randomize some parameters as in~\cite{rebecq2019high} to increase variability in the dataset and reduce the domain gap between synthetic and real data. The contrast threshold, which defines the desired change in brightness to generate an event, is difficult to precisely estimate on real sensors~\cite{gallego2017event} and is instead sampled from a gaussian distribution $\text{N}(0.18, 0.03)$ (where $\text{N}(a, b)$ denotes a gaussian distribution with mean $a$ and standard deviation $b$). Subsequently, the proportion of ambient lighting versus diffuse lighting for the OpenGL rendering engine (employed in the simulator) is randomly sampled from $\text{U}(0, 1)$.
To simulate tracking errors, the center of the bounding box is offset by a random displacement of magnitude $\text{N}(0, 25)$ pixels. Finally, we notice the appearance of white noise captured by the DAVIS346. To quantify the noise, we capture a sequence of a static scene (which should generate no event) with the DAVIS346 and count the number of noisy events generated in each 33~ms window. A gaussian distribution is then fit to the number of noisy events. At training time, we sample a number $k$ from the fitted distribution and randomly select $k$ elements in the set (across $t$, $x$, and $y$) to add uniformly to the input volume. This process is done separately for each polarity (positive and negative). Fig.~\ref{fig:simul_compare} shows the qualitative similarity between real samples acquired with the DAVIS346 and our synthetic samples at the same pose.

\begin{figure}[!t]
    \centering
    \includegraphics[width=\linewidth]{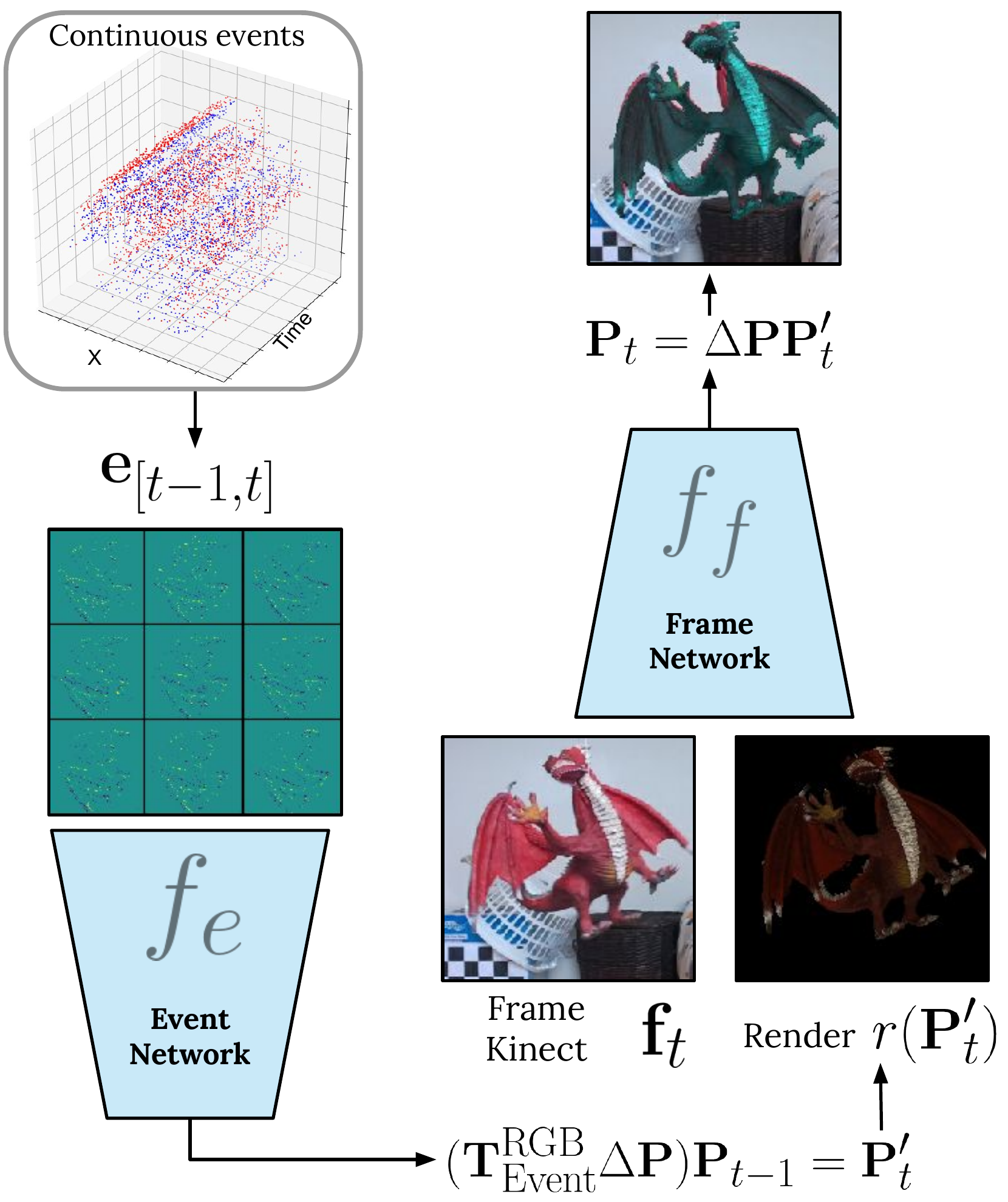}
    \caption{Overview of our method for high-speed tracking with both sensors. On the left, our event network predicts the relative pose changes $\Delta \mathbf{P}$ from $\mathbf{e}_{[t-1,t]}$. The predicted pose is transformed to the RGB referential to estimate the current pose $\mathbf{P}_{t}^\prime$. On the right side, the frame-based network~\cite{garon2018framework} uses the improved pose $\mathbf{P}_{t}^\prime$ to compute a final pose refinement.}
    \label{fig:overview}
\end{figure}

\subsection{Approach overview}

In this paper, we assume that the pose of the object in the previous frame, $\mathbf{P}_{t-1}$, is known. In a full system, it could be initialized by a 3D object detector (e.g. SSD-6D~\cite{kehl2017ssd}) at the first frame ($t=0$). The task of a temporal object tracker is to determine the relative pose \emph{change} $\Delta\mathbf{P}$ between two frames such that an estimate of the current pose $\mathbf{P}_t$ can be obtained by
\begin{equation}
\mathbf{P}_t = \Delta\mathbf{P} \, \mathbf{P}_{t-1} \,.
\end{equation}
Note that all poses $\mathbf{P}_i$ are expressed in the RGB camera coordinate system. 

In this work, we rely on two deep networks to estimate $\Delta\mathbf{P}$. 
First, our novel event network $f_e(\mathbf{e}_{[t-1,t]})$ that takes event data $\mathbf{e}_{[t-1,t]}$ accumulated during the $[t-1, t]$ time interval, and cropped according to the previous object pose $\mathbf{T}_\mathrm{RGB}^\mathrm{Event}\mathbf{P}_{t-1}$. Here, $\mathbf{T}_\mathrm{RGB}^\mathrm{Event} = (\mathbf{T}_\mathrm{Event}^\mathrm{Depth})^{-1} \mathbf{T}_\mathrm{RGB}^\mathrm{Depth}$ is the extrinsic camera calibration matrix from sec.~\ref{sec:spatial_calibration}, necessary to transform the pose estimate in the event camera coordinate system. 

Second, we also employ the RGB-D frame network of Garon et al.~\cite{garon2018framework} $f_f(\mathbf{f}_t, \mathbf{P}_{t-1})$.
Although more recent techniques exist, this choice was made because it is the only one providing both training and inference code and offers robust performance. Alternatively, other approaches such as \cite{joseph2015versatile,Manhardt_2018,marougkas2020track} could also be considered. However, \cite{joseph2015versatile} is already outperformed by \cite{garon2018framework}, the inference code of \cite{Manhardt_2018} is limited to LineMOD objects~\cite{hinterstoisser2012model}, and \cite{marougkas2020track} is an improvement over \cite{garon2018framework} for occlusion handling, which is not the focus here. Note that our method is not limited to this specific network and could be extended to any RGB or RGB-D frame based tracker.

Each network aims to estimate the relative 6-DOF pose of the object.
Interestingly, while events are much more robust to fast displacement they carry less textural information than RGB-D data and we found that the event network used on its own is slightly less accurate.
Therefore, we use a cascade approach where the event network first estimate $\mathbf{P}^\prime_t$, and subsequently the frame network is provided with this new estimation for refinement:
\begin{equation}
\mathbf{P}_t^\prime = (\mathbf{T}^\mathrm{RGB}_\mathrm{Event}f_e(\mathbf{e}_{[t-1,t]})) \, \mathbf{P}_{t-1} \,,
\label{eqn:pose-event}
\end{equation}
\begin{equation}
\mathbf{P}_t = f_f(\mathbf{f}_t, r(\mathbf{P}_t^\prime)) \, \mathbf{P}_t^\prime \,,
\label{eqn:pose-hybrid}
\end{equation}
with $\mathbf{T}_\mathrm{Event}^\mathrm{RGB}$ obtained from the extrinsic camera calibration matrices from sec.~\ref{sec:spatial_calibration} as before. Note that $f_f()$ is an iterative method and can be run multiple time to refine its prediction. To simplify the notation we show a single iteration, in practice, 3 iterations are used as in the original implementation. A diagram overview of the method is provided in fig.~\ref{fig:overview}.

\begin{figure}[!t]
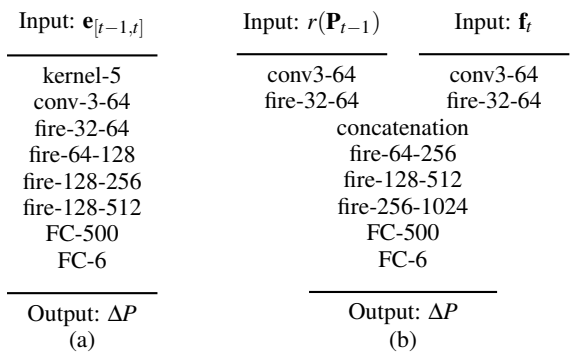

    \centering
    \begin{tabular}{cc}
        \begin{tabular}{c}
            Input: $\mathbf{e}_{[t-1,t]}$ \\
            \noindent\rule{2cm}{.8pt} \\
            kernel-5   \\
            conv-3-64    \\
            fire-32-64   \\
            fire-64-128  \\
            fire-128-256  \\
            fire-128-512 \\
            FC-500       \\
            FC-6         \\
            \noindent\rule{2cm}{.8pt} \\
            Output: $\Delta P$    
        \end{tabular} &
        
        \begin{tabular}{c}
        \begin{tabular}{cc}
            Input: $r(\mathbf{P}_{t-1})$ & Input: $\mathbf{f}_t$ \\
            \noindent\rule{2cm}{.8pt} & \noindent\rule{2cm}{.8pt} \\
            conv3-64 & conv3-64 \\
            fire-32-64 & fire-32-64
        \end{tabular} \\
        \begin{tabular}{c}
            concatenation \\
            fire-64-256 \\
            fire-128-512 \\
            fire-256-1024 \\
            FC-500    \\
            FC-6	 \\
            \noindent\rule{2.5cm}{.8pt} \\
            Output: $\Delta P$  \\
        \end{tabular}
        \end{tabular}
    \\
    (a) & (b)
    \\
    \end{tabular}
    \caption{Our deep network architecture for the (a) event and (b) RGB-D frames. We use the same network architecture as \cite{garon2018framework}. The main differences between both networks are that in (a) we only have one head and a learnable kernel is added to merge temporal information event input data $\mathbf{e}_{[t-1,t]}~$\cite{Gehrig_2019}. The notation ``kernel-$x$'' represents a learnable kernel of dimensions $x$ convoluted on the temporal dimensions with the weights shared for every pixel. The notation ``conv-$x$-$y$'' represent a 2D convolution layer $y$ filters of size $x \times x$, ``fire-x-y'' are ``fire'' modules \cite{iandola2016squeezenet} reducing the channels to $x$ and expanding to $y$ and ``FC-$x$'' are fully connected network of size $x$. Each fire module has skip-links followed by a $2 \times 2$ max pooling. Dropout of 30\% is used after the activation function of each ``fire'' module and both ``FC-500''. All layers, (except ``FC-6'') are followed by an activation function.}
    \label{fig:architecture}
\end{figure}

\subsection{Event network}

Event data is fundamentally different than frame-based data as it possesses two extra dimensions for time and polarity ($T \times P \times X \times Y$, where $T$ is discretized time and $P$ is polarity.). We use the ``Event Spike Tensor'' representation from~\cite{Gehrig_2019} where the time dimension is binned (in our case 9 bins for a 33 ms sample), and the polarity dimension is removed by simply subtracting the negative events from the positive ones.  
Finally, the spatial dimensions are resized as explained in the previous section. The final tensor has a shape of $9 \times 150 \times 150$ where each voxel represents the number of events recorded per time bin. We normalize that quantity between 0 and 1 by dividing each voxel by the maximum amount of events seen in a single voxel during training.

\textbf{Event network architecture.} While the event spike tensor can be processed by a standard CNN, we follow~\cite{Gehrig_2019} and first learn a 1D filter in the time dimension and then apply a standard image convolution where the time dimension acts as different channels. 
In practice, we use the same backbone from~\cite{garon2018framework} for the RGB-D frame network and event network but change only the first two input layers to match the event spike tensor.
Fig.~\ref{fig:architecture} (a) shows the event network architecture. 
The event network is optimized with ADAM~\cite{kingma2014adam} at a learning rate of 0.001 and a batch size of 256. We train for 40 epoch and apply a learning rate scheduling by multiplying the latter by 0.3 every 8 epochs.

\subsection{RGB-D network}
The RGB-D network (see \cite{garon2018framework} for more details) takes as input the current RGB-D frame cropped according to the previous pose $\mathbf{f}_t$ and a rendering of the object at the previous pose $r(\mathbf{P}_{t-1})$. Both inputs have a shape of $4 \times 184 \times 184$ and are normalized by subtracting the mean and dividing by the standard deviation of a subset of the training dataset.
The last layer outputs the predicted 6-DOF pose difference between both inputs.

\textbf{RGB-D network architecture.}
As shown in fig.~\ref{fig:architecture}-(b), each input is individually convoluted then passed to a ``fire'' module~\cite{iandola2016squeezenet}. The module outputs are then concatenate before being max pooled. The single feature map is fed to multiple ``fire'' modules before being applied to two fully connected layers. The RGB-D network is trained with the same optimizer, hyper-parameters and data augmentations from the original work (see \cite{garon2018framework} for more details).

\subsection{Processing Time}
The average inference time is 29.02~ms split in 25.06~ms for the RGB-D network and 3.96~ms for the event network. Note that inference can be reduced close to 25~ms total by running networks in parallel since the total memory footprint is 152.48 MB (frame) + 54.96 MB (event) = 207.44 MB (total), which easily fits on a modern GPU. Runtimes are averaged over 100 samples and computed on an Intel i5 and Nvidia GeForce GTX 1060. 

The numbers reported above include all prepossessing steps such as calculating the bounding box from the last known position and rendering the image for ~\cite{garon2018framework}. Note that building the ``Event Spike Tensor'' representation can be achieved in real time while the previous frame is being processed by the networks (average of 9ms). While our current, unoptimized implementation computes those steps serially, this operation could be trivially parallelized.

%% file: section/05_Evaluation.tex
\section{Experiments}

We now proceed to evaluate our RGB-D-E system for the high-speed tracking of 3D objects in 6-DOF. We first describe our real test dataset, then present quantitative and qualitative results.

\begin{figure}[!t]
	\centering
	\setlength{\tabcolsep}{1pt}
	\begin{tabular}{cc}
	\includegraphics[width=0.48\linewidth]{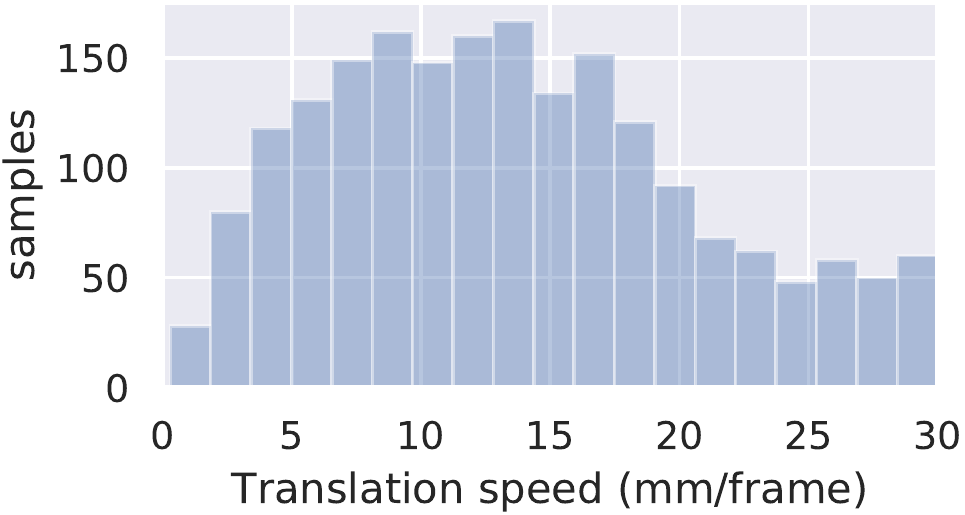} &
	\includegraphics[width=0.48\linewidth]{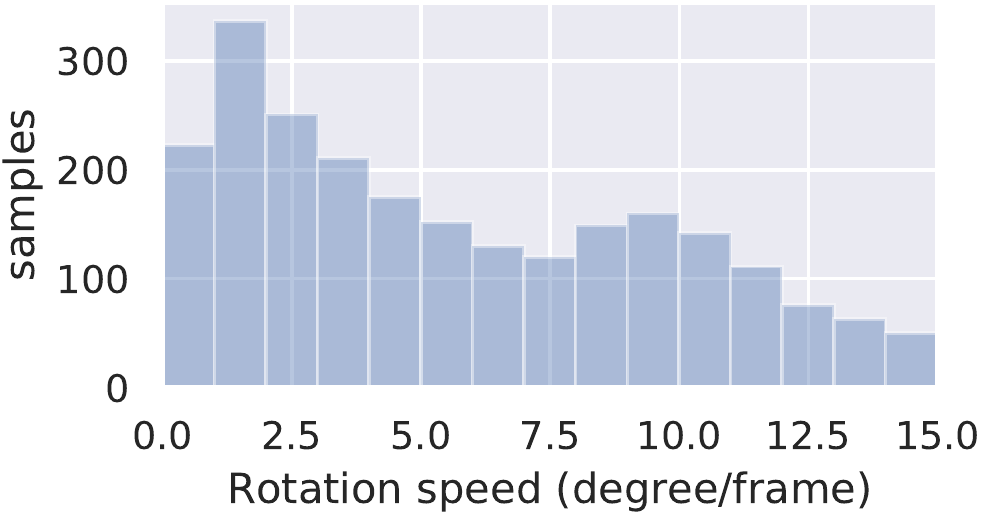} \\
	(a) &
	(b) \\
	\includegraphics[width=0.48\linewidth]{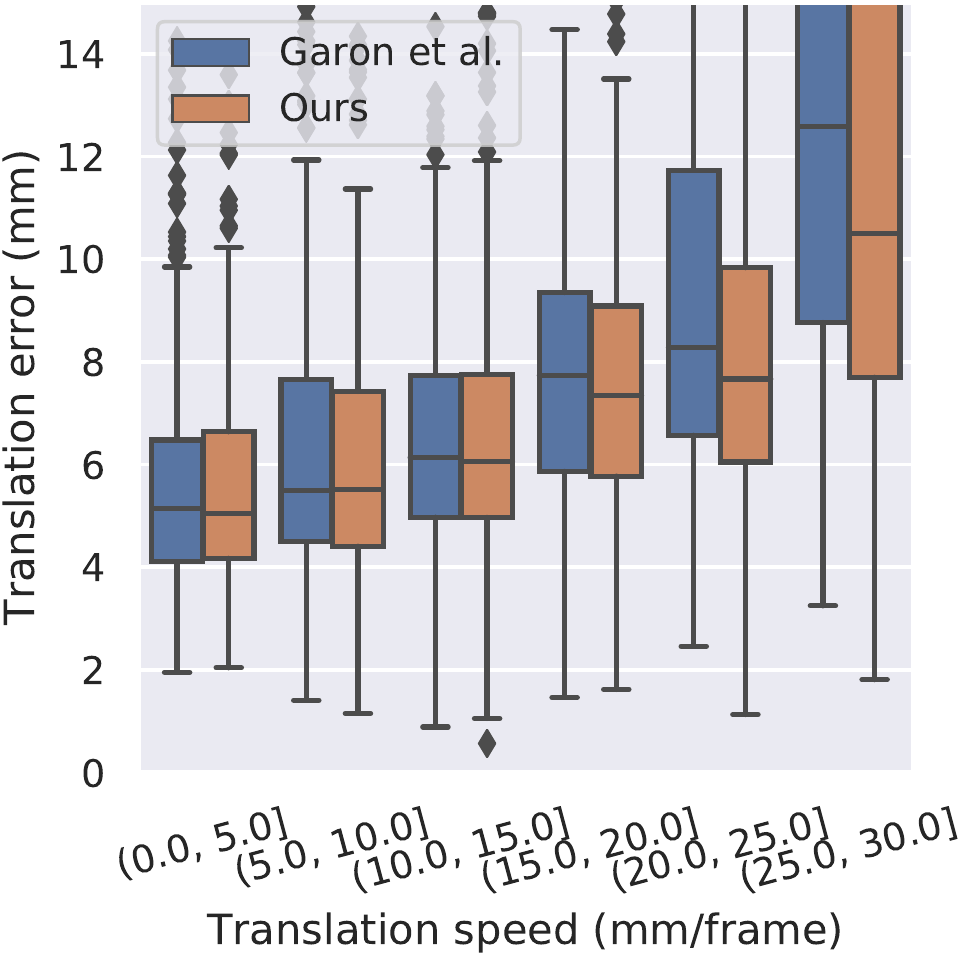} &
	\includegraphics[width=0.48\linewidth]{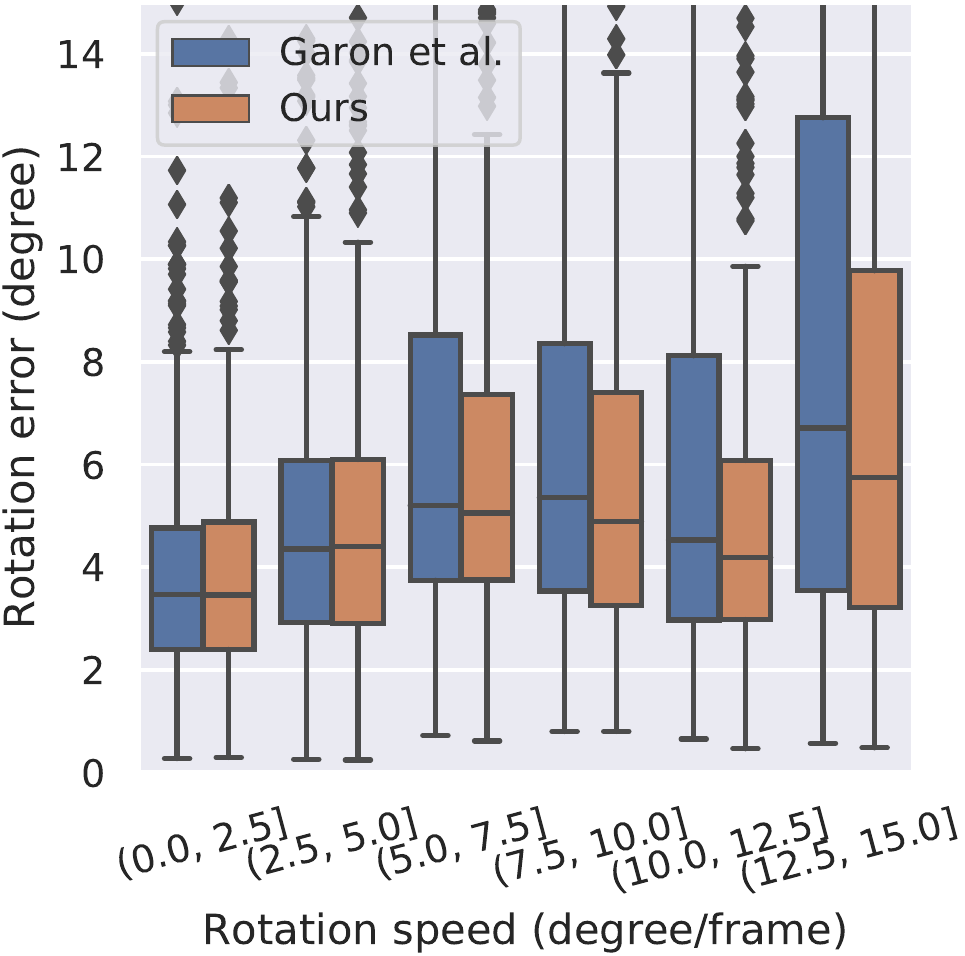} \\
	(c) &
	(d) \\
	\end{tabular}
	\caption{Translation and rotation error as a function of object displacement speed, computed over two consecutive frames. The first row shows the test set distribution indicating the number of frames where the object has a particular (a) translation and (b) rotation speed. The second row plots the distribution of errors between the prediction and the ground truth, computed separately for (c) translation (eq.~\ref{eqn:trans-error}) and (d) rotation (eq.~\ref{eqn:rot-error}). Errors are computed on 10 sequences with a total of 2,472 frames.}
	\label{fig:tracking_errors}
\end{figure}

\begin{figure*}[!t]
\centering
\footnotesize
\setlength{\tabcolsep}{1pt}
\begin{tabular}{cccccccc}

\rotatebox{90}{\hspace{2.5em}Event Frame} & 
\includegraphics[width=.15\linewidth]{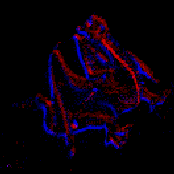} &
\includegraphics[width=.15\linewidth]{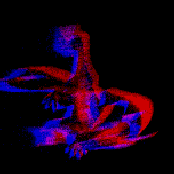} &
\includegraphics[width=.15\linewidth]{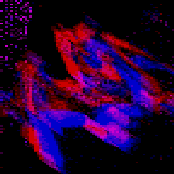} &
\includegraphics[width=.15\linewidth]{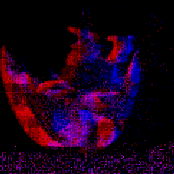} &
\includegraphics[width=.15\linewidth]{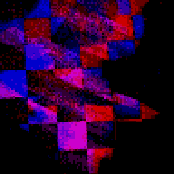} &
\includegraphics[width=.15\linewidth]{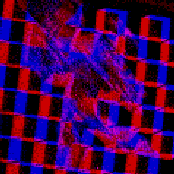} & \\
\rotatebox{90}{\hspace{2.2em}Timestamps} & 
\includegraphics[width=.15\linewidth]{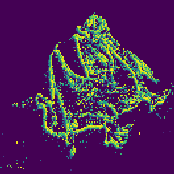} &
\includegraphics[width=.15\linewidth]{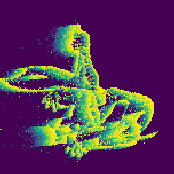} &
\includegraphics[width=.15\linewidth]{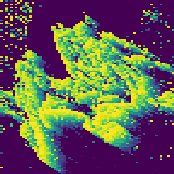} &
\includegraphics[width=.15\linewidth]{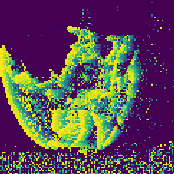} &
\includegraphics[width=.15\linewidth]{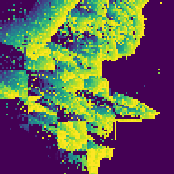} &
\includegraphics[width=.15\linewidth]{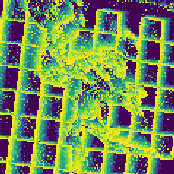} &
\includegraphics[width=.033\linewidth]{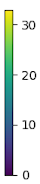} \\
\rotatebox{90}{\hspace{2.5em}RGB} & 
\includegraphics[width=.15\linewidth]{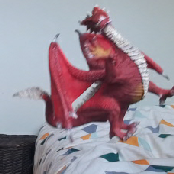} &
\includegraphics[width=.15\linewidth]{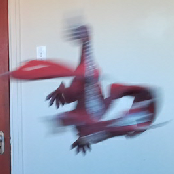} &
\includegraphics[width=.15\linewidth]{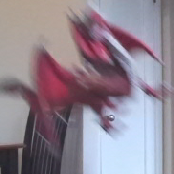} &
\includegraphics[width=.15\linewidth]{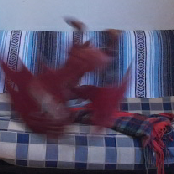} &
\includegraphics[width=.15\linewidth]{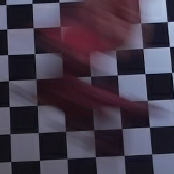} &
\includegraphics[width=.15\linewidth]{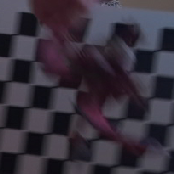} & \\
\rotatebox{90}{\hspace{2.5em}Depth} & 
\includegraphics[width=.15\linewidth]{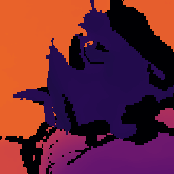} &
\includegraphics[width=.15\linewidth]{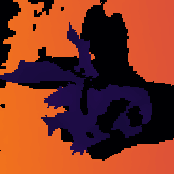} &
\includegraphics[width=.15\linewidth]{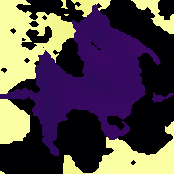} &
\includegraphics[width=.15\linewidth]{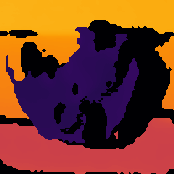} &
\includegraphics[width=.15\linewidth]{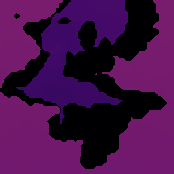} &
\includegraphics[width=.15\linewidth]{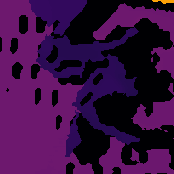} &
\includegraphics[width=.029\linewidth]{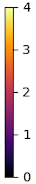} \\
\end{tabular}
\caption{Visualization of the RGB-D-E dataset. Event amplitude (top) from 0 to 5 events accumulate for 33~ms. Event polarities are displayed as blue (positive) and red (negative). Timestamp (second row) associated to each event. Only the last timestamp is displayed for each pixel and range from 0 to 33~ms. Synchronized RGB (third row) and depth map frame (bottom) from the Kinect. The depth map interval is from 0 to 4~m.}
\label{fig:show_dataset}
\end{figure*}

\subsection{Test dataset}

In order to compare the RGB-D and the RGB-D-E trackers, we capture a series of real sequences of a rigid object freely moving at various speeds with different environment perturbation and record the corresponding RGB-D frames and events using our capture setup. To provide a quantitative evaluation, we obtain ground truth pose of the object at each frame using the approach described below. We capture a total of 10~sequences with an average duration of 10~seconds, for a total of 2,472 frames and corresponding event data. Examples from different sequences of the dataset are shown in Fig.~\ref{fig:show_dataset}. The full dataset is publicly available.

For each sequence, we first manually align the 3D model with the object on the first RGB frame. Then, we use ICP~\cite{Pomerleau12comp} to align the visible 3D model vertices with the depth from the RGB-D frame, back-projected in 3D. To avoid back-projecting the entire depth frame, only a bounding box of $(280~\mathrm{mm})^3$ centered around the initial pose is kept. Vertex visibility is computed using raytracing and updated at each iteration of ICP. If the angular pose difference between two successive iterations of ICP is less than $10^\circ$, it is deemed to have converged and that pose is kept. If that condition is not met after a maximum of 10 iterations, ICP diverges and the final pose is refined manually. For all subsequent frames in the sequence, ICP is initialized with the pose from the previous frame. In all, every frame in our test dataset is manually inspected to ensure a good quality pose is obtained, even when it has been determined automatically.

\subsection{Evaluation}

We quantitatively compare our RGB-D-E tracker with the RGB-D approach of Garon et al.~\cite{garon2018framework}, which is the current state-of-the-art in 6-DOF object tracking. We represent a pose $\mathbf{P} = \left[\mathbf{R} \; \mathbf{t}\right]$ by a rotation matrix $\mathbf{R}$ and a translation vector $\mathbf{t}$. The translation error $\delta_\mathbf{t}$ between a pose estimate and its ground truth (denoted by $^*$) is reported as the L2 norm between the two translation vectors
\begin{equation}
\delta_\mathbf{t}(\mathbf{t}^*,\mathbf{t}) = ||\mathbf{t}^* - \mathbf{t}||_2 \,.
\label{eqn:trans-error}
\end{equation}
The rotation error between the two rotation matrices is computed using
\begin{equation}
\delta_\mathbf{R}(\mathbf{R}^*,\mathbf{R}) = \arccos \left( \frac{\mathrm{Tr}(\mathbf{R}^T\mathbf{R}^*) - 1}{2} \right) \,,
\label{eqn:rot-error}
\end{equation}
where $\mathrm{Tr}(\cdot)$ denotes the matrix trace. 

Fig.~\ref{fig:tracking_errors} compares the translation and rotation errors obtained by both approaches. These plots report the error between two adjacent frames only: the trackers are initialized to their ground truth pose at the initial frame. Our method reports lower errors at translation speeds higher than 20~mm/frame, which corresponds to approximately 600~mm/s, and similar rotation errors overall. This is not surprising, given the fact that our method relies on the RGB-D network of Garon et al.~\cite{garon2018framework} to obtain its final pose estimate.

However, visualizing the per-frame error does not tell the whole story. Indeed, in a practical scenario the trackers estimate a succession of predictions instead of being reset to the ground truth pose at every frame. Errors, even small, may therefore accumulate over time and result in tracking failure. Following \cite{garon2018framework}, we consider a tracking failure when either $\delta_\mathbf{t}(\mathbf{t}_i^*, \mathbf{t}_i) > 3~\mathrm{cm}$ or $\delta_\mathbf{R}(\mathbf{R}_i^*,\mathbf{R}_i) > 20^\circ$. Results of this analysis are presented in Tab.~\ref{tab:failure}. The experiment is done at 30fps, 15fps, and 10fps, which is obtained by down-sampling the input frame-based frequency. To allow comparison, the event network is run at the same frequency with the same time window to accumulate the events of 33~ms. For all frame rates, our RGB-D-E approach has at least $61\%$ fewer failures than the RGB-D approach of Garon et al.~\cite{garon2018framework}.

\begin{figure*}[!t]
\centering
\footnotesize
\setlength{\tabcolsep}{1pt}
\begin{tabular}{ccccccccc}

\rotatebox{90}{\hspace{.0em}Garon et al.~\cite{garon2018framework}} & 
\includegraphics[width=.11\linewidth]{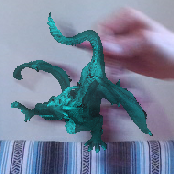} &
\includegraphics[width=.11\linewidth]{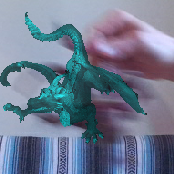} &
\includegraphics[width=.11\linewidth]{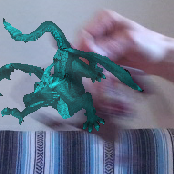} &
\includegraphics[width=.11\linewidth]{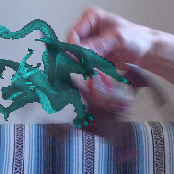} &
\includegraphics[width=.11\linewidth]{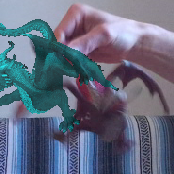} &
\includegraphics[width=.11\linewidth]{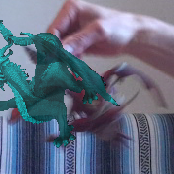} &
\includegraphics[width=.11\linewidth]{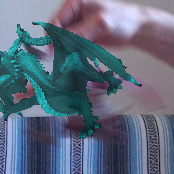} &
\includegraphics[width=.11\linewidth]{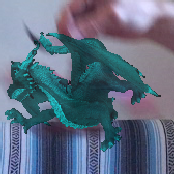} \\
\rotatebox{90}{\hspace{1.6em}Ours} & 
\includegraphics[width=.11\linewidth]{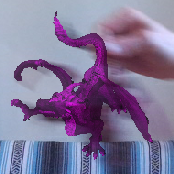} &
\includegraphics[width=.11\linewidth]{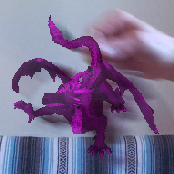} &
\includegraphics[width=.11\linewidth]{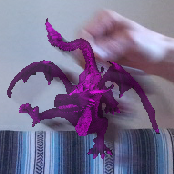} &
\includegraphics[width=.11\linewidth]{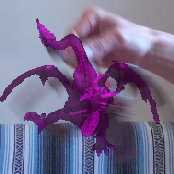} &
\includegraphics[width=.11\linewidth]{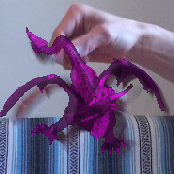} &
\includegraphics[width=.11\linewidth]{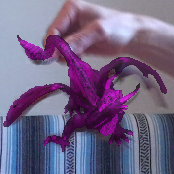} &
\includegraphics[width=.11\linewidth]{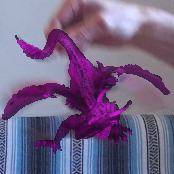} &
\includegraphics[width=.11\linewidth]{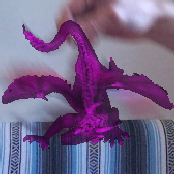} \\
\rotatebox{90}{\hspace{0.5em}Ground truth} & 
\includegraphics[width=.11\linewidth]{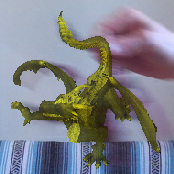} &
\includegraphics[width=.11\linewidth]{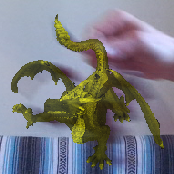} &
\includegraphics[width=.11\linewidth]{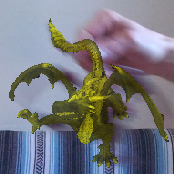} &
\includegraphics[width=.11\linewidth]{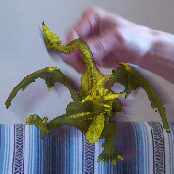} &
\includegraphics[width=.11\linewidth]{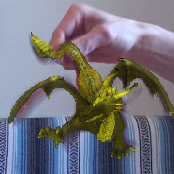} &
\includegraphics[width=.11\linewidth]{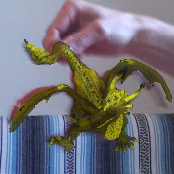} &
\includegraphics[width=.11\linewidth]{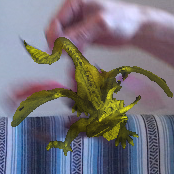} &
\includegraphics[width=.11\linewidth]{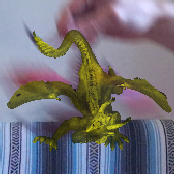} \\

&
&
&
&
&
&
&
&
\\

\rotatebox{90}{\hspace{.0em}Garon et al.~\cite{garon2018framework}} & 
\includegraphics[width=.11\linewidth]{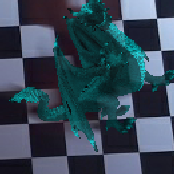} &
\includegraphics[width=.11\linewidth]{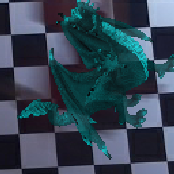} &
\includegraphics[width=.11\linewidth]{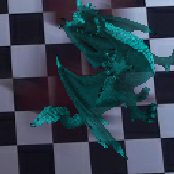} &
\includegraphics[width=.11\linewidth]{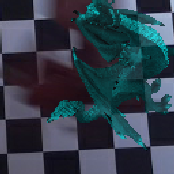} &
\includegraphics[width=.11\linewidth]{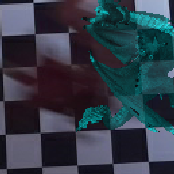} &
\includegraphics[width=.11\linewidth]{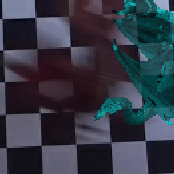} &
\includegraphics[width=.11\linewidth]{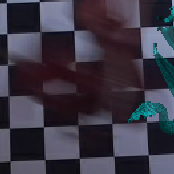} &
\includegraphics[width=.11\linewidth]{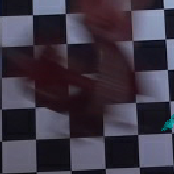} \\
\rotatebox{90}{\hspace{1.6em}Ours} & 
\includegraphics[width=.11\linewidth]{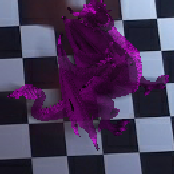} &
\includegraphics[width=.11\linewidth]{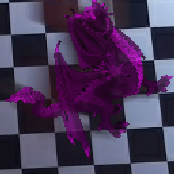} &
\includegraphics[width=.11\linewidth]{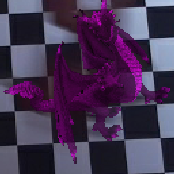} &
\includegraphics[width=.11\linewidth]{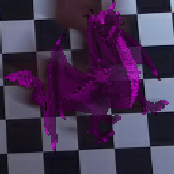} &
\includegraphics[width=.11\linewidth]{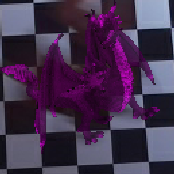} &
\includegraphics[width=.11\linewidth]{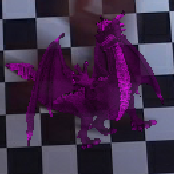} &
\includegraphics[width=.11\linewidth]{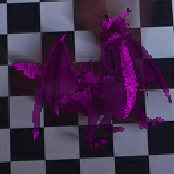} &
\includegraphics[width=.11\linewidth]{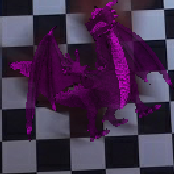} \\
\rotatebox{90}{\hspace{0.5em}Ground truth} & 
\includegraphics[width=.11\linewidth]{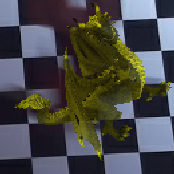} &
\includegraphics[width=.11\linewidth]{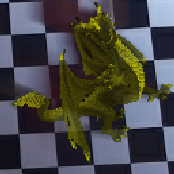} &
\includegraphics[width=.11\linewidth]{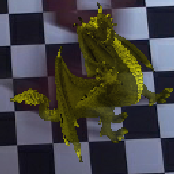} &
\includegraphics[width=.11\linewidth]{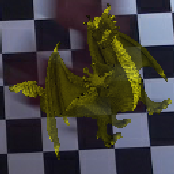} &
\includegraphics[width=.11\linewidth]{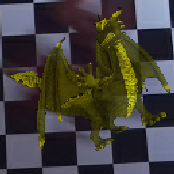} &
\includegraphics[width=.11\linewidth]{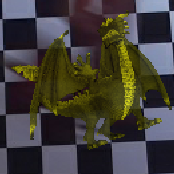} &
\includegraphics[width=.11\linewidth]{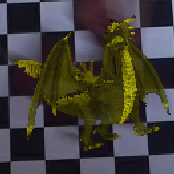} &
\includegraphics[width=.11\linewidth]{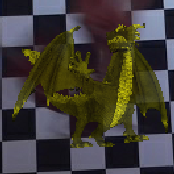} \\
\end{tabular}
\caption{Qualitative comparison between the different approaches on different sequences. The overlay is the position predicted by each method. From top to bottom, Garon et al.~\cite{garon2018framework} (blue), ours (pink) and the ground truth (yellow). Each frame is continuous in the sequence and cropped according to the ground truth position.}
\label{fig:seq}
\end{figure*}

Fig.~\ref{fig:seq} shows representative qualitative results comparing both techniques with the ground truth. Those results show that the approach of Garon et al.~\cite{garon2018framework} is affected by the strong motion blur which arises under fast object motion. In contrast, our approach remains stable and can follow the object through very fast motion. \textbf{Please see video results in the supplementary materials}.

\begin{table}
\centering
\begin{tabular}{lccc}
\toprule
\textbf{Method}  &  \multicolumn{3}{c}{\textbf{Failures}} \\ 
   & 30fps & 15fps &  10fps \\ 
\midrule
Garon et al.~\cite{garon2018framework} & 83 & 130 & 166 \\
Ours   & 28  & 48 & 64 \\
\bottomrule
\end{tabular}
\caption{Number of tracking failures for each method with multiple RGB-D camera's frame rates. Failures are computed on 10 independent sequences with a total of 2,472 frames.}
\label{tab:failure}
\end{table}

%% file: section/06_Discussion.tex
\section{Discussion}

We present a novel acquisition setup for simultaneous RGB-D-E capture which combines a Kinect Azure camera with a DAVIS346 sensor. With the new event modality, we show that a state-of-the-art RGB-D 6-DOF object tracker can be significantly improved in terms of tracking speed. We capture an evaluation dataset with ground truth 3D object poses that mimics difficult scenarios typically encountered in augmented reality applications : a user manipulating a small object with fast free motions. Using this dataset, we demonstrate that our approach achieves a threefold decrease in loss of tracking over the previous state-of-the-art, thereby bringing 6-DOF object tracking closer to applicability in real-life scenarios. 

\textbf{Limitations and future work.} First, capturing an evaluation dataset is time-consuming and obtaining the 6-DOF ground truth pose of the object is difficult, especially when fast motions are involved. While our semi-automatic approach provided a way to acquire a small number of sequences easily, scaling up to larger RGB-D-E datasets will require more sophisticated apparatus such as a motion capture (mocap) setup as in \cite{garon2018framework}. Indeed, mocap systems are ideal for this use case as they can track the object robustly at high frame rates. Second, while using a cascade scheme improves significantly the robustness to large motion of the tracker, it is still inherently limited in accuracy since it always relies on the frame network. The success of the cascade configuration motivates further exploration of better ways to fuse the Event modality with the previous frame-based modalities. Third, we notice that the trackers are still sensitive to dynamic backgrounds (see the last example in the supplementary video). We anticipate that this could be partially solved by generating training data with spurious structured events such as those that could be created by a dynamic background (or a moving camera). These represent exciting future research directions that we plan to investigate in order to achieve even more robust and accurate object tracking systems that can be used in real-world augmented reality applications.